\documentclass[10pt,twocolumn,letterpaper]{article}

\usepackage{iccv}
\usepackage{times}
\usepackage{epsfig}
\usepackage{graphicx}
\usepackage{amsmath}
\usepackage{amssymb}
\usepackage{booktabs}
\usepackage{paralist}
\usepackage{subcaption}
\usepackage{soul}
\usepackage{cite}
\usepackage[pagebackref=true,breaklinks=true,letterpaper=true,colorlinks,bookmarks=false]{hyperref}

\DeclareGraphicsExtensions{.pdf,.jpg,.png}
\graphicspath{{figs/}}

\newcommand{\secref}[1]{Section~\ref{sec:#1}}
\newcommand{\figref}[1]{Figure~\ref{fig:#1}}
\newcommand{\tblref}[1]{Table~\ref{tbl:#1}}

\iccvfinalcopy 

\def\iccvPaperID{1102} 

\ificcvfinal\fi

\begin{document}

\title{A Unified Model for Near and Remote Sensing}

\author{
  \begin{minipage}{\linewidth}
    \centering
    \begin{minipage}{1.5in}
      \centering
      Scott Workman$^1$\\
      {\tt\small scott@cs.uky.edu}
    \end{minipage}
    \begin{minipage}{1.5in}
      \centering
      Menghua Zhai$^1$\\
      {\tt\small ted@cs.uky.edu}
    \end{minipage}
    \begin{minipage}{1.5in}
      \centering
      David J. Crandall$^2$\\
      {\tt\small djcran@indiana.edu}
    \end{minipage}
    \begin{minipage}{1.5in}
      \centering
      Nathan Jacobs$^1$\\
      {\tt\small jacobs@cs.uky.edu}
    \end{minipage}
    \\[.2cm]
    \hspace{-14pt}
    \begin{minipage}{1.7in}
      $^1$University of Kentucky
    \end{minipage}
    \begin{minipage}{2.2in}
      $^2$Indiana University Bloomington
    \end{minipage}
  \end{minipage}
}

\maketitle

\begin{abstract}

  We propose a novel convolutional neural network architecture for
  estimating geospatial functions such as
  population density, land cover, or land use. In our approach, we
  combine overhead and ground-level images in an end-to-end trainable
  neural network, which uses kernel regression and density
  estimation to convert features extracted from the ground-level
  images into a dense feature map. The output of this network is a
  dense estimate of the geospatial function in the form of a
  pixel-level labeling of the overhead image. To evaluate our
  approach, we created a large dataset of overhead and ground-level
  images from a major urban area with three sets of labels: land use,
  building function, and building age. We find that our approach is
  more accurate for all tasks, in some cases dramatically so.

\end{abstract}

\section{Introduction}

From predicting the weather to planning the future of our cities to
recovering from natural disasters, accurately monitoring widespread
areas of the Earth's surface is essential to many scientific
fields and to society in general. These observations have traditionally been
collected through remote sensing from satellites, aerial imaging,
and distributed observing stations and sensors.  These approaches can
observe certain properties like land cover and land use
accurately and at a high resolution, but  unfortunately, not
everything can be seen from overhead imagery. For example, Wang et
al.~\cite{wang2016torontocity} evaluate approaches for urban zoning
and building height estimation from overhead imagery, and conclude
that urban zoning segmentation ``is an extremely hard task from aerial
views,'' that building height estimation is ``either too hard, or more
sophisticated methods are needed,'' and that ``utilizing ground
imagery seems a logical first step.''

\begin{figure}

  \centering

  \includegraphics[width=.95\linewidth]{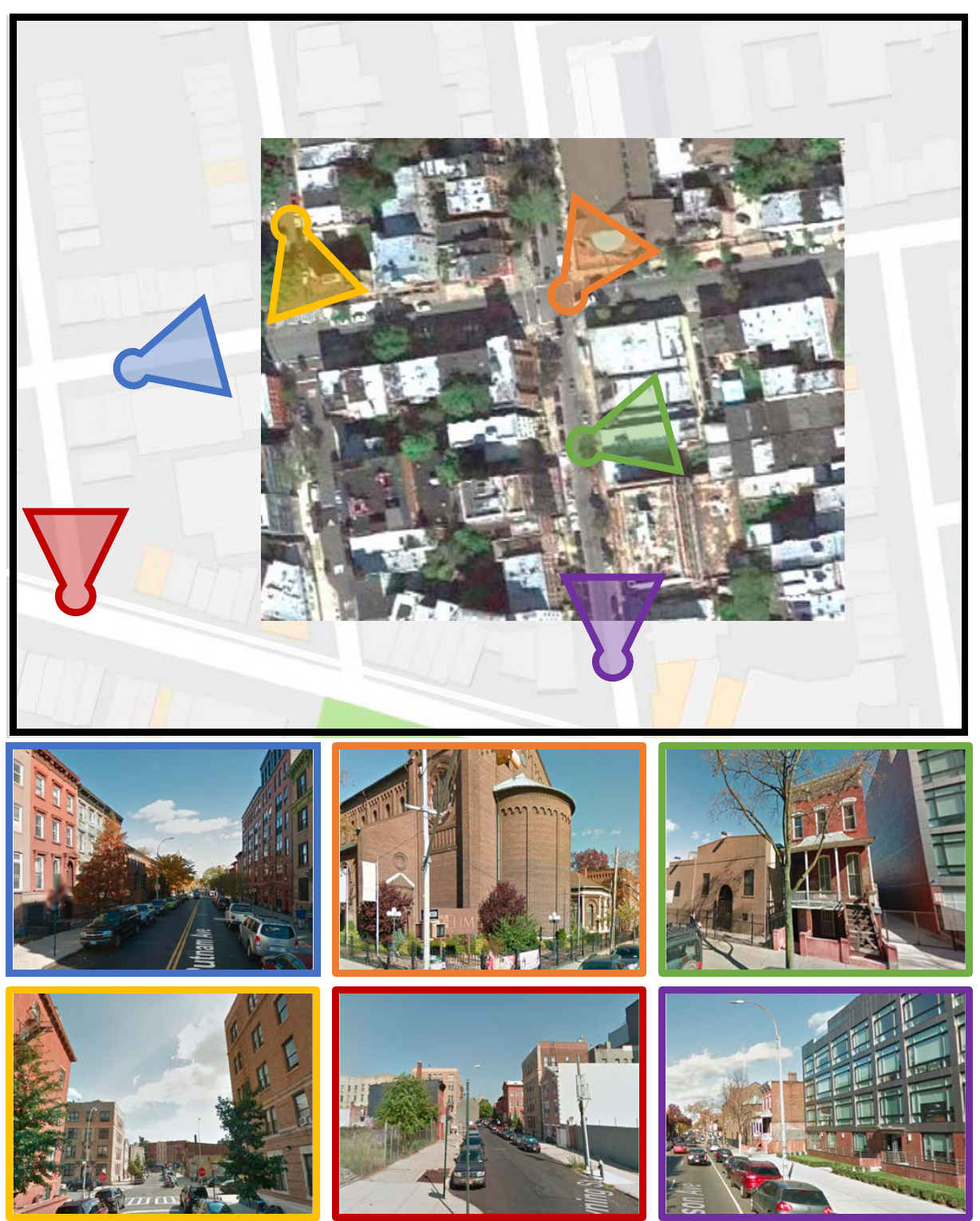} 

  \caption{We use overhead imagery and geotagged ground-level imagery
  as input to an end-to-end deep network that estimates the values of a
  geospatial function by performing fine-grained pixel-level labeling
  on the overhead image.}

  \label{fig:cartoon}

\end{figure}

More recently, the explosive popularity of geotagged social media has
raised the possibility of using online user-generated content as a
source of geospatial information, sometimes called {\em image-driven
mapping} or {\em proximate sensing}. For example, online images from
social network and photo sharing websites have been used to
estimate land cover for large geographic
regions~\cite{leung2010proximate,zhu2015land}, to observe the state of
the natural world by recreating maps of snowfall~\cite{wang2016mm},
and to quantify perception of urban
environments~\cite{dubey2016deep}. Despite differing applications,
these papers all wish to estimate some unobservable {\em
geospatial function}, and view each social media artifact (\eg,
geotagged ground-level image) as an observation of this function at a
particular geographic location.  

The typical
approach~\cite{arietta2014city,xie2011im2map} is
to (1) collect a large number of samples, (2) use an automated
approach to estimate the value of the geospatial function for each
sample, and (3) use some form of locally weighted averaging to
interpolate the sparse samples into a dense, coherent estimate of the
underlying geospatial function. This estimation is complicated by the
fact that observations are noisy; state-of-the-art recognition
algorithms are imperfect, some images are inherently confusing or
ambiguous, and the observations are distributed sparsely and
non-uniformly. This means that in order to estimate geospatial
functions with reasonable accuracy, most techniques use a kernel with
a large bandwidth to smooth out the noise, which yields coarse,
low-resolution outputs.  Despite this limitation, the proximate
sensing approach can work well if ground-level imagery is plentiful,
the property is easily estimated from the imagery, and the geospatial
function is smoothly varying.  

We propose a novel neural network architecture that combines the
strengths of these two approaches (\figref{cartoon}). Our approach uses deep
convolutional neural networks (CNNs) to extract features from both
overhead and ground-level imagery. For the ground-level images, we use
kernel regression and density estimation to convert the sparsely
distributed feature samples into a dense feature map spatially
consistent with the overhead image. This differs from the proximate
sensing approach, which uses kernel regression to directly estimate
the geospatial function. Then, we fuse the ground-level feature map
with a hidden layer of the overhead image CNN. To extend our methods
to pixel-level labeling, we extract multiscale features in the form of
a hypercolumn and use a small neural network to estimate the
geospatial function of interest. A novel element of our approach is
the use of a spatially varying kernel that depends on features extracted
from the overhead imagery.

\begin{figure}

  \centering

  \begin{subfigure}{1\linewidth}
    \centering
    \includegraphics[height=.265\linewidth]{{{motivation/40.673500_-73.961658_fire_station}}}
    \includegraphics[height=.265\linewidth]{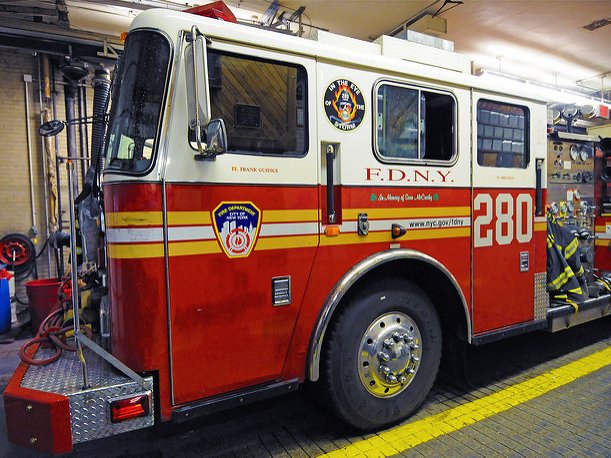}
    \includegraphics[height=.265\linewidth]{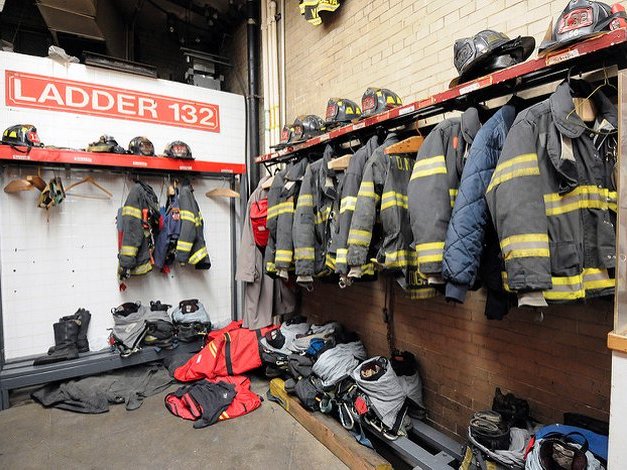}
  \end{subfigure}

  \caption{What type of building is shown in the overhead view
  (left)? Identifying and mapping building function is a challenging
  task that becomes considerably easier when taking into context
  nearby ground-level imagery (right).}

  \label{fig:motivation}

\end{figure}

Our network is trained end-to-end, so that all free parameters,
including kernel bandwidths and low-level image features, are
automatically tuned to minimize our loss function. In addition, our
architecture is very general because it could be used with most
state-of-the-art CNNs, and could be easily adapted to use any sparsely
distributed media, including geotagged audio, video, and text (\eg,
tweets). We evaluate our approach with a large real-world dataset,
consisting of most of two major boroughs of New York City (Brooklyn
and Queens), on estimating three challenging labels (building age,
building function, and land use), all of which are notoriously
challenging tasks in remote sensing (\figref{motivation}).  The
results show that our technique for fusing overhead and ground-level
imagery is more accurate than either the remote or proximate sensing
approach alone, and that our automatically-estimated spatially-varying
kernel improves accuracy compared to one that is uniform. The dataset
and our implementation will be made available at our project
website.\footnote{\url{http://cs.uky.edu/~scott/research/unified/}}

\section{Related Work}

\begin{figure*}[t]

  \centering

  \includegraphics[width=1\linewidth]{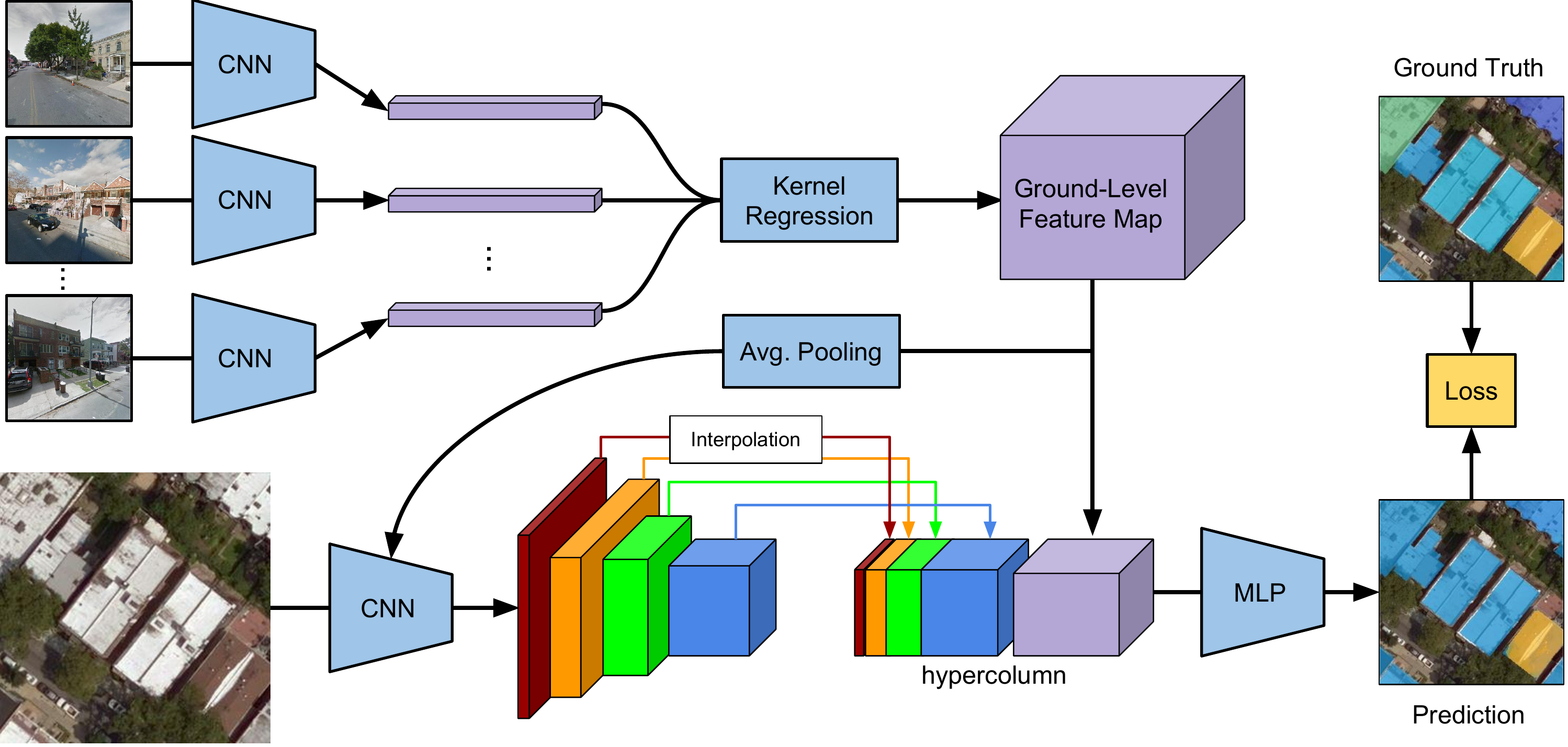} 

  \caption{An overview of our network architecture.}

\label{fig:architecture}

\end{figure*}

Many recent studies have explored analyzing large-scale
image collections as a means of characterizing properties of the
physical world.  A number of papers have tried to
estimate properties of weather from geotagged and time-stamped
ground-level imagery. For example, Murdock et
al.~\cite{murdock2015building, murdock2013webcam2satellite} and Jacobs et al.~\cite{jacobs2016cloudmap} use webcams
to infer cloud cover maps, Li et al.~\cite{li2015smog} use
ground-level photos to estimate smog conditions, Glasner et
al.~\cite{glasner2015hot} estimate temperature, Zhou et
al.~\cite{zhou2014recognizing} and Lee et al.~\cite{lee2014predicting}
estimate demographic properties, Fedorov et
al.~\cite{fedorov2015snowwatch, fedorov2014snow} and Wang et
al.~\cite{wang2016mm} infer snow cover, Khosla et
al.~\cite{khosla2014looking} and Porzi et
al.~\cite{porzi2015predicting} measure perceived crime levels, Leung
and Newsam~\cite{leung2010proximate} estimate land use, and so on.

Many of these papers' contribution is exploring a novel application,
as opposed to proposing novel techniques.  They mostly follow a very
similar recipe in which standard recognition techniques are applied to
individual images, and then spatial smoothing and other noise
reduction techniques are used to create an estimate of the geospatial
function across the world. Meanwhile, remote sensing has long used
computer vision to estimate properties of the Earth from satellite
images.  Of course, overhead imaging is quite different from
ground-level imaging, and so remote sensing techniques have largely
been developed independently and in task-specific ways~\cite{Rozen}.

We know of relatively little work that has proposed general frameworks
for estimating geospatial functions from imagery, or in
integrating visual evidence from both ground-level and overhead image
viewpoints. Tang et al.~\cite{tang2015improving} show how location
context can improve image classification, but they do not use
overhead imagery and their goal is not to estimate geospatial
functions.  Luo et al.~\cite{luo2008event} use overhead imagery to
give context for event recognition in ground-level photos
by combining hand-crafted features for each modality. Xie et
al.~\cite{xie2015transfer} use transfer learning  to extract
socioeconomic indicators from overhead imagery.
Most similar is our work on mapping the subjective attribute of
natural beauty~\cite{workman2017beauty} where we propose to use a
multilayer perceptron to combine high-level semantic features.
Recent work in image
geolocalization has matched ground-level photos taken at
unknown locations to georegistered overhead
views~\cite{lin2013cross,lin2015learning,workman2015geocnn,workman2015wide},
but this goal is significantly different from inferring geospatial
functions of the world.

Several recent papers jointly reason about co-located ground-level and overhead
image pairs. M{\'a}ttyus et al.~\cite{mattyus2016hd} perform joint
inference over both monocular aerial and ground-level images from a
stereo camera for fine-grained road segmentation, while  Wegner
et al.~\cite{wegner2016cataloging} detect and classify
trees using features extracted from overhead and ground-level images.
Ghouaiel and Lef{\`e}vre~\cite{ghouaiel2016coupling} transform
ground-level panoramas to an overhead perspective for change 
detection. Zhai et al.~\cite{zhai2017predicting}
propose 
a transformation to extract meaningful features from overhead
imagery.

In contrast with the above work, our goal is to produce a general
framework for learning that can estimate any given geospatial
function of the world. We integrate data from both ground-level
imagery, which often contains visual evidence that is not visible from
the air, and overhead imagery, which is typically much denser. We
demonstrate how our models learn in an end-to-end way, avoiding the
need for task-specific or hand-engineered features.

\section{Problem Statement}

We address the problem of estimating a spatially varying property of
the physical world, which we model as an unobservable mathematical
function that maps latitude-longitude coordinates to possible values
of the property, $F: \mathbb{R}^2 \rightarrow {\cal Y}$. The range
${\cal Y}$ of this function depends on the attribute to be estimated,
and might be categorical (\eg, a discrete set of elements for land use
classification --- golf course, residential, agricultural, \etc) or
continuous (\eg, population density).  We wish to estimate this
function based on the available observable evidence, including data
sampled both densely (such as overhead imagery) and sparsely (such as
geotagged ground-level images).  From a probabilistic
perspective, we can think of our task as learning a conditional
probability distribution $P(F(l)=y | S_l, \mathbf{G}(l))$, where $l$ is a latitude-longitude coordinate, $S_l$ is an overhead image centered at
that location, and $\mathbf{G}(l)$ is a set of nearby ground-level images.

\section{Network Architecture}

We propose a novel convolutional neural network (CNN) 
that fuses high-resolution overhead imagery and nearby ground-level
imagery to estimate the value of a geospatial function at a target location. While
we focus on images, our overall architecture could be used with
many sources of dense and sparse data. Our network can be trained in
an end-to-end manner, which enables it to learn to optimally
extract features from both the dense and sparse data sources. 

\subsection{Architecture Overview}

The overall architecture of our network (\figref{architecture})
consists of three main components, the details of which we describe in
the next several sections: (1) constructing a spatially dense feature
map using features extracted from the ground-level images
(\secref{ground}), (2) extracting features from the overhead image,
incorporating the ground-level image feature map (\secref{overhead}),
and (3) predicting the geospatial function value based on a
hypercolumn of features (\secref{prediction}).   A novel element of
our proposed approach is the use of an adaptive, spatially varying
interpolation method for constructing the ground-level image feature
map based on features extracted from the overhead image
(\secref{adaptive}).  

\subsection{Ground-Level Feature Map Construction}
\label{sec:ground}

The goal of this component is to convert a sparsely sampled set of
ground-level images into a dense feature map. For a given geographic
location $l$, let $\mathbf{G}(l) = \{(G_{i},{l_i})\}$ be a set of $N$
elements corresponding to the closest ground-level images, where each $(G_i, l_i)$ is an image and its respective geographic
location. We use a CNN to extract features, $f_g(G_i)$, from
each image and interpolate using Nadaraya--Watson kernel regression,
\begin{equation}
  f_G(l) = \frac{\sum w_i f_g(G_i)}{\sum w_i},
  \label{eq:lwa}
\end{equation}
where $w_i=exp(-d(l,l_i;\Sigma)^2)$ is a Gaussian kernel function
where a diagonal covariance matrix $\Sigma$ controls the kernel
bandwidth and $d(l,l_i;\Sigma)$ is the Mahalanobis distance from $l$
to $l_i$.  We perform this interpolation for every pixel location in
the overhead image. The result is a feature map of size $H \times W \times
m$, where $H$ and $W$ are the height and width of the overhead image
in pixels, and $m$ is the output dimensionality of our ground-level
image CNN.  

The diagonal elements of the covariance matrix are represented by a
pair of trainable weights, which pass through a {\em softplus}
function (\ie $f(x) = ln(1+e^x)$) to ensure they are positive. Here,
the value of $\Sigma$ does not depend on geographic location, a
strategy we call {\em uniform}. In \secref{adaptive}, we propose an
approach in which $\Sigma$ is spatially varying.

In our experiments, the ground-level images, $\mathbf{G}(l)$, are
actually geo-oriented street-level panoramas. To form a
feature representation for each panorama, $G_i$, we first extract
perspective images in the cardinal directions, resulting in four
ground-level images per location. We replicate the ground-level image
CNN, $f_g(G_i)$, four times, feed each image through separately, and
concatenate the individual outputs. We then add a final $1\times1$
convolution to reduce the feature dimensionality. For our experiments,
we use the VGG-16 architecture~\cite{simonyan2014very}, initialized with weights for Place
categorization~\cite{zhou2017places} ($m=205$, layer name `fc8').
The result is an 820 dimensional feature vector for each location,
which is further reduced to 50 dimensions.

It is possible that the nearest ground-level image may be far away, 
which could lead to later processing stages  incorrectly
interpreting the feature map. To overcome this, we concatenate a
kernel density estimate, using the kernel defined in equation~(\ref{eq:lwa}), of the ground-level
image locations to the ground-level image feature map.  The result is
an $H \times W \times 51$ feature map that captures appearance and
distributional information of the ground-level images.

\begin{figure*}
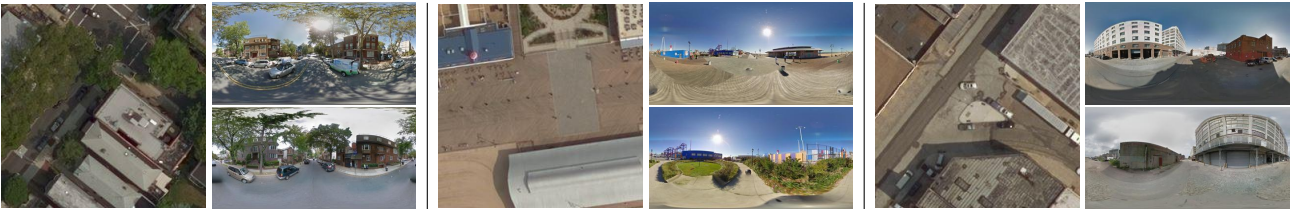


  \centering

  \begin{subfigure}{.155\linewidth}
    \centering
    \includegraphics[height=1\linewidth]{{{dataset/154397_197283}}}
  \end{subfigure}
  \begin{subfigure}{.155\linewidth}
    \centering
    \includegraphics[height=.5\linewidth]{{{dataset/154397_197283_1}}}
    \includegraphics[height=.5\linewidth]{{{dataset/154397_197283_4}}}
  \end{subfigure}
\,\vrule\,
  \begin{subfigure}{.155\linewidth}
    \centering
    \includegraphics[height=1\linewidth]{{{dataset/154401_197390}}}
  \end{subfigure}
  \begin{subfigure}{.155\linewidth}
    \centering
    \includegraphics[height=.5\linewidth]{{{dataset/154401_197390_1}}}
    \includegraphics[height=.5\linewidth]{{{dataset/154401_197390_5}}}
  \end{subfigure}
\,\vrule\,
  \begin{subfigure}{.155\linewidth}
    \centering
    \includegraphics[height=1\linewidth]{{{dataset/154353_197232}}}
  \end{subfigure}
  \begin{subfigure}{.155\linewidth}
    \centering
    \includegraphics[height=.5\linewidth]{{{dataset/154353_197232_2}}}
    \includegraphics[height=.5\linewidth]{{{dataset/154353_197232_6}}}
  \end{subfigure}

  \caption{Sample overhead imagery and nearby street-level panoramas
  included in the Brooklyn and Queens dataset.}

\label{fig:dataset}

\end{figure*}

\subsection{Overhead Feature Map Construction}
\label{sec:overhead}

This section describes the CNN we use to extract features from the
overhead image and how we integrate the ground-level feature map.  The
CNN is based on the VGG-16 architecture~\cite{simonyan2014very}, which
has 13 convolutional layers, each using $3 \times 3$ convolutions, and
three fully connected layers. We only use the convolutional layers,
typically referred to as conv-$\{1_{1-2}, 2_{1-2}, 3_{1-3}, 4_{1-3},
5_{1-3}\}$. In addition, we reduce the dimensionality of the feature
maps that are output by each layer. These layers have output dimensionality
of $\{32, 64, 128, 256, 512\}$ channels, respectively. Each
intermediate layer uses a leaky ReLU activation function
($\alpha=0.2$).

To fuse the ground-level feature map with the overhead imagery, we
apply average pooling with a kernel size of $6 \times 6$ and a stride
of 2. Given an input overhead image with $H=W=256$, this
reduces the ground-level feature map to $32 \times 32 \times 51$. We
then concatenate it, in the channels dimension, with the overhead
image feature map at the seventh convolutional layer, $3_3$.  The
input to convolutional layer $4_1$ is then $32 \times 32 \times 179$.
We experimented with including the ground-level feature map earlier
and later in the network and found this to be a good tradeoff between
computational cost and expressiveness.

\subsection{Geospatial Function Prediction}
\label{sec:prediction}

Given an overhead image, $S_l$, we use the ground-level and overhead
feature maps defined above as input to the final component of our
system to estimate the value of the geospatial function,
$F(l(p))\in{1\ldots K}$, where $l(p)$ is the location of a pixel $p$.
This pixel might be the center of the image for the image
classification setting or any arbitrary pixel in the pixel-level
labeling setting. To accomplish this we adapt ideas from the PixelNet
architecture~\cite{bansal2017pixelnet}, due to its strong performance
and ability to train using sparse inputs. However, our approach for
incorporating sparsely distributed inputs could be adapted to other
semantic labeling architectures.

We first resize each feature map to be $H\times W$ using bilinear
interpolation.  We then extract a {\em hypercolumn}~\cite{hariharan2015hypercolumns} consisting of a
set of features centered around $p$, $h_p(S) = [c_1(S,p), c_2(S,p),
  \ldots, ...  c_M(S,p)]$, where $c_i$ is the feature map of the
$i$-th layer.  For this work, we extract hypercolumn features from
conv-$\{1_2, 2_2, 3_3, 4_3, 5_3\}$ and the ground-level feature map.
The resulting hypercolumn feature has length 1,043. Note that
resizing all intermediate feature maps to be the size of the image is
quite memory intensive. Following Bansal et
al.~\cite{bansal2017pixelnet}, we subsample pixels during training to
increase the number (and therefore diversity) of images per
mini-batch. At testing time, we can either compute the hypercolumn for
all pixels to create a dense semantic labeling or a subset to label
particular locations.

This hypercolumn feature is then passed to a small multilayer
perceptron (MLP) that provides the estimate of the geospatial
function. The MLP has three layers of size 512, 512, and $K$ (the
task dependent number of outputs).  Each intermediate layer uses a
leaky ReLU activation function.

\subsection{Adaptive Kernel Bandwidth Estimation}
\label{sec:adaptive}

In addition to the {\em uniform} kernel described above for forming
the ground-level image feature map (\secref{ground}), we propose an
{\em adaptive} strategy that predicts the optimal kernel bandwidth
parameters for each location in the feature map.  We estimate these
bandwidth parameters using a CNN applied to the overhead image. This
network shares the first three groups of convolutional layers,
conv-$\{1_1,\ldots,3_3\}$, with the overhead image CNN defined in
\secref{overhead}. The output of these convolutions is passed to a
sequence of three convolutional transpose layers, each with filter
size $3 \times 3$ and a stride of 2. These layers have output
dimensionality of 32, 16, and 2, respectively.  The final layer has an
output size of $H \times W \times 2$, which represents the diagonal
entries of the kernel bandwidth matrix, $\Sigma$, for each pixel
location. Similar to the {\em uniform} approach, we apply a {\em
softplus} activation on the output (initialized with a small constant
bias) to ensure positive kernel bandwidth. When using the {\em
adaptive} strategy, these bandwidth parameters are used to construct the
ground-level feature map ($H \times W \times 51$). 

\section{Experiments}

We evaluated the performance of our approach on a challenging
real-world dataset, which includes overhead imagery, ground-level
imagery, and several fine-grained pixel-level labels.  We proposed two
variants of our approach: {\em unified (uniform)}, which uses a single
kernel bandwidth for the entire region, and {\em unified (adaptive)},
which uses a location-dependent kernel that is conditioned on the
overhead image.

\subsection{Baseline Methods}

In order to  evaluate the proposed macro-architecture, we
use several baseline methods that share many low-level components
with our proposed methods. 
\begin{compactitem}

  \item {\em random} represents random sampling from the
    prior distribution of the training dataset.

  \item {\em remote}  represents the traditional remote
    sensing approach, in which only overhead imagery is used. We use
    the {\em unified (uniform)} architecture, but do not incorporate
    the ground-level feature map in the overhead image CNN or the
    hypercolumn.

  \item {\em proximate}  represents the proximate sensing
    approach in which only ground-level imagery is used. We start from
    the {\em unified (uniform)} architecture but only include the
    ground-level image feature map (minus the kernel density estimate)
    in the hypercolumn.

  \item {\em grid}  is similar to the {\em proximate}
    method. Starting from {\em unified (uniform)}, we omit
    all layers from the overhead image CNN prior to concatenating in
    the ground-level feature map from the hypercolumn. The motivation
    for this method is that the additional convolutional layers are
    able to capture spatial patterns which the final MLP cannot,
    because it operates on individual hypercolumns.

\end{compactitem}

\subsection{Implementation Details}

All methods were implemented using Google's TensorFlow
framework~\cite{abadi2016tensorflow} and optimized using
ADAM~\cite{kingma2014adam} with default training parameters, except
for an initial learning rate of $10^{-3}$ (decreasing by $0.5$ every
7,500 mini-batches) and weight decay of $5\times 10^{-4}$. During
training, we randomly sampled 2,000 pixels per image per
mini-batch. The ground-level CNNs have shared weights. All other
network weights were randomly initialized and allowed to vary freely.
We applied batch normalization~\cite{ioffe2015batch} (decay $=0.99$)
in all convolutional and fully connected layers (except for output
layers). For our experiments, we minimize a cross entropy loss
function and consider the nearest 20 street-level panoramas.  Each
network was trained for 25 epochs with a batch size of 32 on an NVIDIA
Tesla P100.

\subsection{Brooklyn and Queens Dataset}

We introduce a new dataset containing ground-level
and overhead images from Brooklyn and Queens, two boroughs of New York
City (\figref{dataset}).  It consists of non-overlapping overhead images downloaded from
Bing Maps (zoom level 19, approximately $30cm$ per pixel) and
street-level panoramas from Google Street View.  From Brooklyn, we
collected imagery for the entirety of King's County.  This resulted in
73,921 overhead images and 139,327 panoramas. A significant
number (30,316) of the overhead images are over water; we discard
these and only consider those which contain buildings.  We hold out
4,361 overhead images for testing.  For Queens, we selected a held
out region solely for evaluation and used the same process to collect
imagery.  This resulted in a dataset with 10,044 overhead images
and 38,603 panoramas. 

Using data made publicly available by NYC Open
Data,\footnote{\url{https://data.cityofnewyork.us/}} we constructed a
per-pixel labeling of each overhead image for the following set of
labels.

\paragraph{Building Function.}
We used 206 building classes, as outlined by the New York City
Department of City Planning (NYCDCP) in the Primary Land Use Tax Lot
Output (PLUTO) dataset, to categorize each building in a given
overhead image.  PLUTO contains detailed geographic data at the tax
lot level (property boundary) for every piece of land in New York
City. Example labels include: Multi-Story Department Stores, Funeral
Home, and Church. To this set we add two classes, background
(non-building, such as roads and water) and unknown, as there are
several thousand unlabeled tax lots. To form our final labeling, we
intersected the tax lot data with building footprints obtained from
the NYC Planimetric Database. For reference, there are approximately
331,000 buildings in Brooklyn.

\paragraph{Land Use.}
From PLUTO, we generated a per-pixel label image with each contained
tax lot labeled according to its primary land use category. The land
use categories were specified by the New York City Department of City
Planning. In total, there are 11 land use categories. Example land use
categories include: One and Two Family Buildings, Commercial and
Office Buildings, and Open Space and Outdoor Recreation. Similar to
building function, we add two classes, background (\eg, roads) and
unknown.

\paragraph{Building Age.}
Again using PLUTO in conjunction with the NYC Planimetric Database,
we generated a per-pixel label image with each building
labeled according to the year that construction of the building was
completed. Brooklyn and Queens have a lengthy history, with the oldest
building on record dating to the mid-1600s. We quantize time by
decades, with a bin for all buildings constructed before 1900. This
resulted in 13 bins, to which we added a bin for background
(non-building), as well as unknown for a small number of buildings
without a documented construction year.

\subsection{Semantic Segmentation}

We report results using pixel accuracy and region intersection over
union averaged over classes (mIOU), two standard metrics for the
semantic segmentation task. In both cases, higher is better. When
computing these metrics, we ignore any ground-truth pixel labeled as
unknown. In addition, for the tasks of building function and age
estimation, we ignore background pixels.

\begin{figure*}
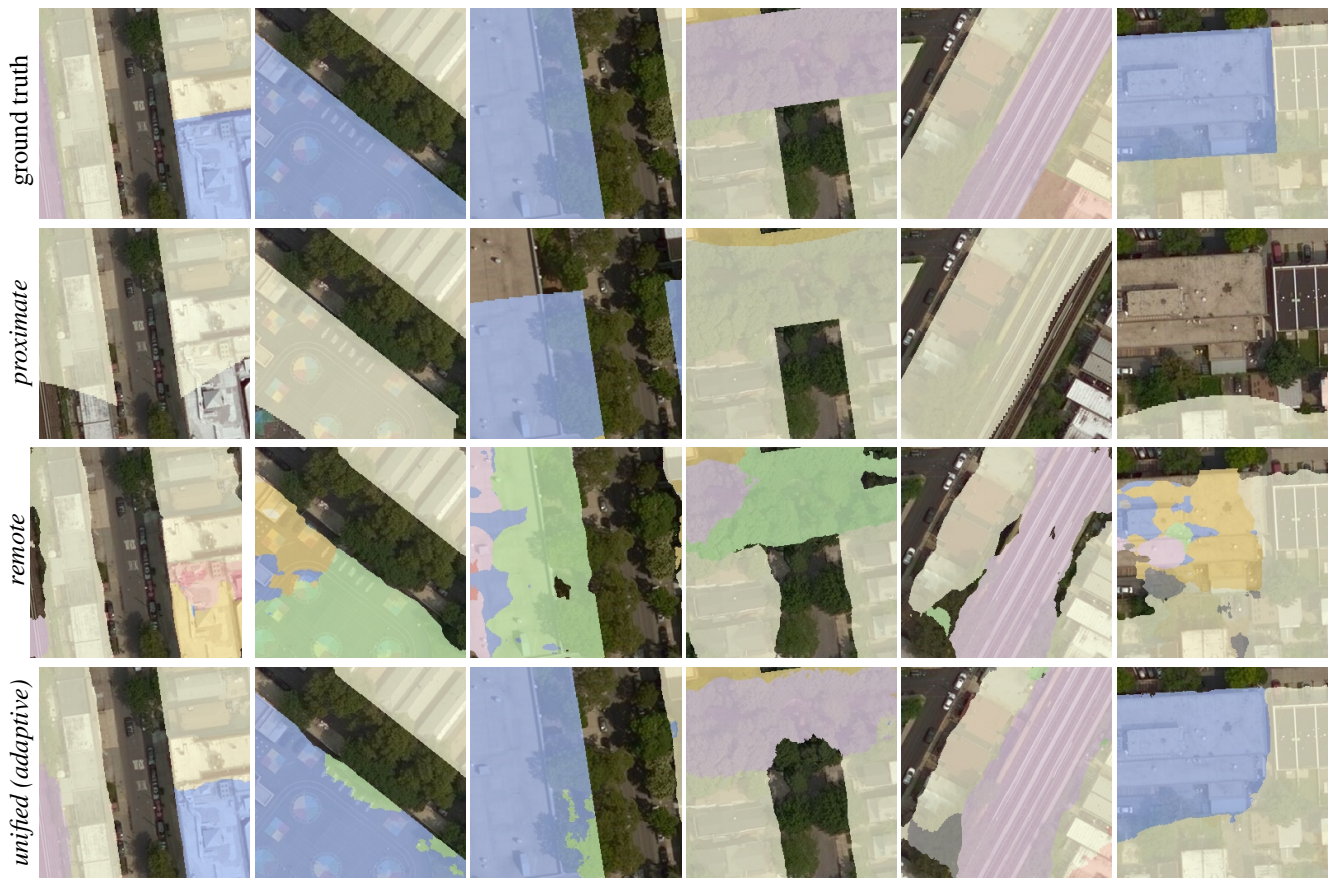


  \centering

  \setlength\tabcolsep{1pt}

    \begin{tabular}{lcccccc}
      \raisebox{.25\height}{\rotatebox{90}{ground truth}}
      \includegraphics[width=.16\linewidth]{{{results/landuse/154436_197326}}} &
      \includegraphics[width=.16\linewidth]{{{results/landuse/154382_197317}}} &
      \includegraphics[width=.16\linewidth]{{{results/landuse/154415_197157}}} &
      \includegraphics[width=.16\linewidth]{{{results/landuse/154420_197284}}} &
      \includegraphics[width=.16\linewidth]{{{results/landuse/154434_197376}}} &
      \includegraphics[width=.16\linewidth]{{{results/landuse/154446_197245}}} \\
      
      \raisebox{.5\height}{\rotatebox{90}{\em proximate}}
      \includegraphics[width=.16\linewidth]{{{results/landuse/154436_197326_lwa}}} &
      \includegraphics[width=.16\linewidth]{{{results/landuse/154382_197317_lwa}}} &
      \includegraphics[width=.16\linewidth]{{{results/landuse/154415_197157_lwa}}} &
      \includegraphics[width=.16\linewidth]{{{results/landuse/154420_197284_lwa}}} &
      \includegraphics[width=.16\linewidth]{{{results/landuse/154434_197376_lwa}}} &
      \includegraphics[width=.16\linewidth]{{{results/landuse/154446_197245_lwa}}} \\
      
      \raisebox{1\height}{\rotatebox{90}{\em remote}}
      \includegraphics[width=.16\linewidth]{{{results/landuse/154436_197326_satellite}}} &
      \includegraphics[width=.16\linewidth]{{{results/landuse/154382_197317_satellite}}} &
      \includegraphics[width=.16\linewidth]{{{results/landuse/154415_197157_satellite}}} &
      \includegraphics[width=.16\linewidth]{{{results/landuse/154420_197284_satellite}}} &
      \includegraphics[width=.16\linewidth]{{{results/landuse/154434_197376_satellite}}} &
      \includegraphics[width=.16\linewidth]{{{results/landuse/154446_197245_satellite}}} \\
      
      \raisebox{.06\height}{\rotatebox{90}{\em unified (adaptive)}}
      \includegraphics[width=.16\linewidth]{{{results/landuse/154436_197326_adaptive}}} &
      \includegraphics[width=.16\linewidth]{{{results/landuse/154382_197317_adaptive}}} &
      \includegraphics[width=.16\linewidth]{{{results/landuse/154415_197157_adaptive}}} &
      \includegraphics[width=.16\linewidth]{{{results/landuse/154420_197284_adaptive}}} &
      \includegraphics[width=.16\linewidth]{{{results/landuse/154434_197376_adaptive}}} &
      \includegraphics[width=.16\linewidth]{{{results/landuse/154446_197245_adaptive}}} \\
    \end{tabular}

  \caption{Sample results for classifying land use: (top--bottom)
  ground truth, {\em proximate}, {\em remote}, and {\em unified
  (adaptive)}.}

  \label{fig:landuse_results}

\end{figure*}

\begin{table}
  \centering
  \caption{Brooklyn evaluation results (top-1 accuracy).}
  \begin{tabular}{@{}lrrr@{}}
    \toprule
    & \multicolumn{1}{c}{Age} & \multicolumn{1}{c}{Function} & \multicolumn{1}{c}{Land Use} \\
    \bottomrule
    {\em random}             &  6.82\% &  0.49\% &  8.55\% \\
    {\em proximate}          & 35.90\% & 27.14\% & 44.66\% \\
    {\em grid}               & 38.68\% & 33.84\% & 71.64\% \\
    {\em remote}             & 37.18\% & 34.64\% & 69.63\% \\
    \hline
    {\em unified (uniform)}  & \textbf{44.08\%} & 43.88\% & 76.14\% \\
    {\em unified (adaptive)} & 43.85\% & \textbf{44.88\%} & \textbf{77.40\%} \\
    \bottomrule
  \end{tabular}
  \label{tbl:brooklyn_top1}
\end{table}

\begin{table}
  \centering
  \caption{Brooklyn evaluation results (mIOU).}
  \begin{tabular}{@{}lrrr@{}}
    \toprule
    & \multicolumn{1}{c}{Age} & \multicolumn{1}{c}{Function} & \multicolumn{1}{c}{Land Use} \\
    \bottomrule
    {\em random}             &  2.76\% & 0.11\% &  3.21\% \\
    {\em proximate}          & 11.77\% & 5.46\% & 18.04\% \\
    {\em grid}               & 16.98\% & 9.37\% & 37.76\% \\
    {\em remote}             & 15.11\% & 4.67\% & 31.70\% \\
    \hline
    {\em unified (uniform)}  & 20.88\% & 13.66\% & 43.53\% \\
    {\em unified (adaptive)} & \textbf{23.13\%} & \textbf{14.59\%} & \textbf{45.54\%} \\
    \bottomrule
  \end{tabular}
  \label{tbl:brooklyn_miou}
\end{table}

\begin{table}
  \centering
  \caption{Queens evaluation results (top-1 accuracy).}
  \begin{tabular}{@{}lrrr@{}}
    \toprule
    & \multicolumn{1}{c}{Age} & \multicolumn{1}{c}{Function} & \multicolumn{1}{c}{Land Use} \\
    \bottomrule
    {\em random}             &  6.80\% &  0.49\% &  8.41\% \\
    {\em proximate}          & 25.27\% & 22.50\% & 47.40\% \\
    {\em grid}               & 27.47\% & 26.62\% & 67.51\% \\
    {\em remote}             & 26.06\% & 29.85\% & 69.27\% \\
    \hline
    {\em unified (uniform)}  & 29.68\% & 33.64\% & 68.08\% \\
    {\em unified (adaptive)} & \textbf{29.76\%} & \textbf{34.13\%} & \textbf{70.55\%} \\
    \bottomrule
  \end{tabular}
  \label{tbl:queens_top1}
\end{table}

\begin{table}
  \centering
  \caption{Queens evaluation results (mIOU).}
  \begin{tabular}{@{}lrrr@{}}
    \toprule
    & \multicolumn{1}{c}{Age} & \multicolumn{1}{c}{Function} & \multicolumn{1}{c}{Land Use} \\
    \bottomrule
    {\em random}             & 2.58\% & 0.09\% &  3.05\% \\
    {\em proximate}          & 5.08\% & 1.57\% & 15.04\% \\
    {\em grid}               & 7.31\% & 2.30\% & 28.02\% \\
    {\em remote}             & 7.78\% & 2.67\% & 28.46\% \\
    \hline
    {\em unified (uniform)}  & 8.95\% & 3.71\% & 31.03\% \\
    {\em unified (adaptive)} & \textbf{9.53\%} & \textbf{3.73\%} & \textbf{33.48\%} \\
    \bottomrule
  \end{tabular}
  \label{tbl:queens_miou}
\end{table}

\begin{figure*}
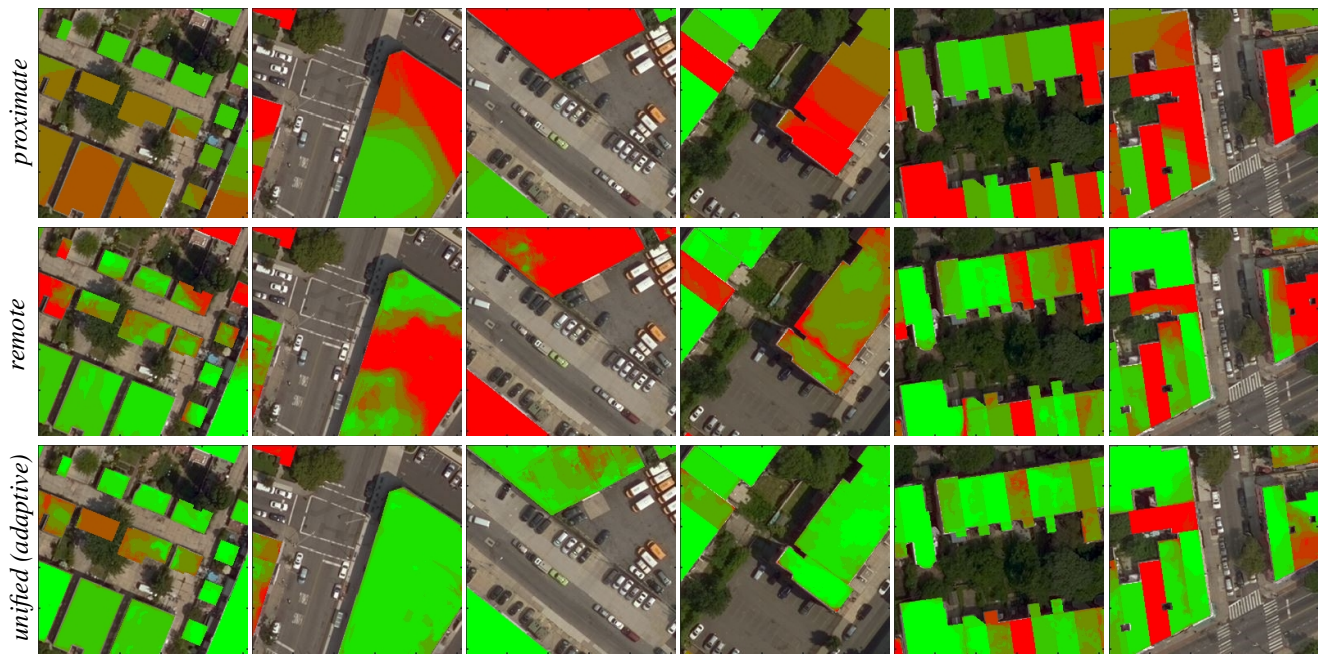


  \centering

  \setlength\tabcolsep{1pt}

  \begin{tabular}{lcccccc}
    \raisebox{.5\height}{\rotatebox{90}{\em proximate}} &
    \includegraphics[width=.159\linewidth]{{{results/function/154326_197265_lwa}}} &
    \includegraphics[width=.159\linewidth]{{{results/function/154326_197300_lwa}}} &
    \includegraphics[width=.159\linewidth]{{{results/function/154397_197258_lwa}}} &
    \includegraphics[width=.159\linewidth]{{{results/function/154511_197165_lwa}}} &
    \includegraphics[width=.159\linewidth]{{{results/function/154449_197180_lwa}}} &
    \includegraphics[width=.159\linewidth]{{{results/function/154419_197247_lwa}}} \\
    
    \raisebox{.9\height}{\rotatebox{90}{\em remote}} &
    \includegraphics[width=.159\linewidth]{{{results/function/154326_197265_satellite}}} &
    \includegraphics[width=.159\linewidth]{{{results/function/154326_197300_satellite}}} &
    \includegraphics[width=.159\linewidth]{{{results/function/154397_197258_satellite}}} &
    \includegraphics[width=.159\linewidth]{{{results/function/154511_197165_satellite}}} &
    \includegraphics[width=.159\linewidth]{{{results/function/154449_197180_satellite}}} &
    \includegraphics[width=.159\linewidth]{{{results/function/154419_197247_satellite}}} \\
    
    \raisebox{.07\height}{\rotatebox{90}{\em unified (adaptive)}} &
    \includegraphics[width=.159\linewidth]{{{results/function/154326_197265_adaptive}}} &
    \includegraphics[width=.159\linewidth]{{{results/function/154326_197300_adaptive}}} &
    \includegraphics[width=.159\linewidth]{{{results/function/154397_197258_adaptive}}} &
    \includegraphics[width=.159\linewidth]{{{results/function/154511_197165_adaptive}}} &
    \includegraphics[width=.159\linewidth]{{{results/function/154449_197180_adaptive}}} &
    \includegraphics[width=.159\linewidth]{{{results/function/154419_197247_adaptive}}} \\
  \end{tabular}

  \caption{Sample results for identifying building function. From
  top to bottom, we visualize top-k images for the {\em proximate},
  {\em remote}, and {\em unified (adaptive)} methods, respectively.
  Each pixel is color coded on a scale from green to red by the rank
  of the correct class in the posterior distribution, where bright
  green is the best (rank one).}

  \label{fig:function_results}

\end{figure*}

\begin{figure}
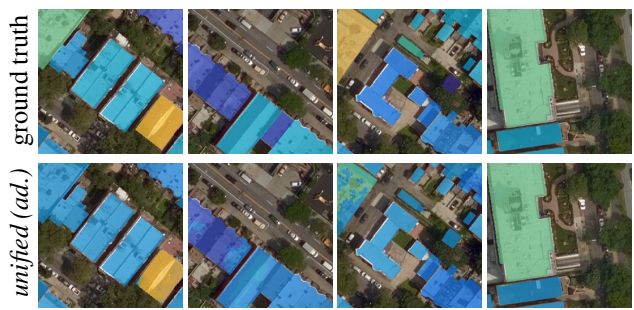


  \centering

  \setlength\tabcolsep{1pt}

  \begin{subfigure}{1\linewidth}
    \begin{tabular}{lcccc}
      \raisebox{.06\height}{\rotatebox{90}{ground truth}} &
      \includegraphics[width=.23\linewidth]{{{results/age/154337_197284}}} &
      \includegraphics[width=.23\linewidth]{{{results/age/154345_197259}}} &
      \includegraphics[width=.23\linewidth]{{{results/age/154345_197283}}} &
      \includegraphics[width=.23\linewidth]{{{results/age/154418_197309}}} \\
      
      \raisebox{.08\height}{\rotatebox{90}{\em unified (ad.)}} &
      \includegraphics[width=.23\linewidth]{{{results/age/154337_197284_adaptive}}} &
      \includegraphics[width=.23\linewidth]{{{results/age/154345_197259_adaptive}}} &
      \includegraphics[width=.23\linewidth]{{{results/age/154345_197283_adaptive}}} &
      \includegraphics[width=.23\linewidth]{{{results/age/154418_197309_adaptive}}} \\
    \end{tabular}
  \end{subfigure}

  \caption{Sample results for estimating building age: (top) ground
  truth and (bottom) {\em unified (adaptive)}.}

  \label{fig:age_results}

\end{figure}

\paragraph{Classifying Land Use.}
We consider the task of identifying a parcel of land's primary land
use. This task is considered especially challenging from an overhead
only perspective, with recent work simplifying the task by considering
only three classes~\cite{wang2016torontocity}. We report top-1
accuracy for land use classification using the Brooklyn test set in
\tblref{brooklyn_top1} and on Queens in \tblref{queens_top1}.
Similarly we report mIOU for Brooklyn and Queens in
\tblref{brooklyn_miou} and \tblref{queens_miou}, respectively. Our
results support the notion that this task is extremely difficult.
However, our approach, {\em unified (adaptive)}, is significantly
better than all baselines, including an overhead image only approach
({\em remote}). Qualitative results for this task are shown in
\figref{landuse_results}.

\paragraph{Identifying Building Function.}
We consider the task of making a functional map of buildings. To our
knowledge, our work is the first to explore this. For example, in
\figref{motivation}, it becomes considerably easier to identify that
the building in the overhead image is a fire station when shown two
nearby ground-level images. We report performance metrics for this
task in \tblref{brooklyn_top1} and \tblref{queens_top1} for accuracy,
and \tblref{brooklyn_miou} and \tblref{queens_miou} for mIOU.
Qualitative results are shown in \figref{function_results}.  Given the
challenging nature of this task, we visualize results as a top-k
image, where each pixel is colored from green (best) to red, by the
rank of the correct class in the posterior distribution. Our approach
produces labelings much more consistent with the ground truth.

\paragraph{Estimating Building Age.}
Finally, we consider the task of estimating the year a building was
constructed. Intuitively, this is an extremely difficult task from an
overhead image only viewpoint, but is also non-trivial from a
ground-level view. We report accuracy and mIOU metrics for this
experiment in \tblref{brooklyn_top1} and \tblref{brooklyn_miou} for
the Brooklyn region and \tblref{queens_top1} and in
\tblref{queens_miou} for Queens. Our approach significantly
outperforms the baselines. Example qualitative results are shown in
\figref{age_results}.

\subsection{Does Known Orientation Help?}

In the evaluation above, we constructed the ground-level feature map
(\secref{ground}) using features from geo-oriented panorama cutouts.
The cutout images were extracted in the cardinal directions and their
features stacked in a fixed order.  To better understand the value of
the ground-level feature map, we investigated how knowing the
orientation of the ground-level images affects accuracy.  We repeated
the land use classification experiment on Brooklyn using our {\em
uniform (adaptive)} approach (retraining the network), but randomly
circular-shifted the set of images prior to feature extraction. Note
that orientation is not completely random, because doing so would have
required regenerating cutouts.  We observe a significant performance
drop from $77.40\%$ to $72.61\%$ in top-1 accuracy, about $3\%$ higher
than using the overhead image only method. This experiment shows that
knowing the orientation of the ground-level images is critical for
achieving the best performance, but that including the ground-level
images without knowing the orientation can still be useful.

\section{Conclusion}

We proposed a novel neural network architecture for estimating
geospatial functions and evaluated it in the context of fine-grained
understanding of an urban area.  Our network fuses overhead and
ground-level images and gives more accurate predictions than if either
modality had been used in isolation.  Specifically, our approach is
better at resolving spatial boundaries than if only ground-level
images were used and is better at estimating features that are
difficult to determine from a purely overhead perspective.  A key
feature of our architecture is that it is end-to-end trainable,
meaning that it can learn to extract the optimal features, for any
appropriate loss function, from the raw pixels of all images, as well
as parameters used to control the fusion process.  While we
demonstrated its use with ground-level images, our architecture is
general and could be used with a wide variety of sparsely distributed
measurements, including geotagged tweets, video, and audio.

\ificcvfinal
\paragraph*{Acknowledgments}

We gratefully acknowledge the support of NSF CAREER grants IIS-1553116
(Jacobs) and IIS-1253549 (Crandall), a Google Faculty Research Award
(Jacobs), and an equipment donation from IBM to the University of
Kentucky Center for Computational Sciences.

\fi

\clearpage
{\small
\bibliographystyle{ieee}
\bibliography{biblio}
}

\newpage
\null
\vskip .375in
\twocolumn[{%
  \begin{center}
    \textbf{\Large Supplemental Material : \\ A Unified Model for Near and Remote Sensing}
  \end{center}
  \vspace*{24pt}
}]
\setcounter{page}{1}
\setcounter{section}{0}
\setcounter{equation}{0}
\setcounter{figure}{0}
\setcounter{table}{0}
\makeatletter
\renewcommand{\theequation}{S\arabic{equation}}
\renewcommand{\thefigure}{S\arabic{figure}}

This document contains additional details and experiments related to
our methods. 

\section{Brooklyn and Queens Dataset}

\begin{figure*}
 
  \centering

  \includegraphics[width=.4\linewidth]{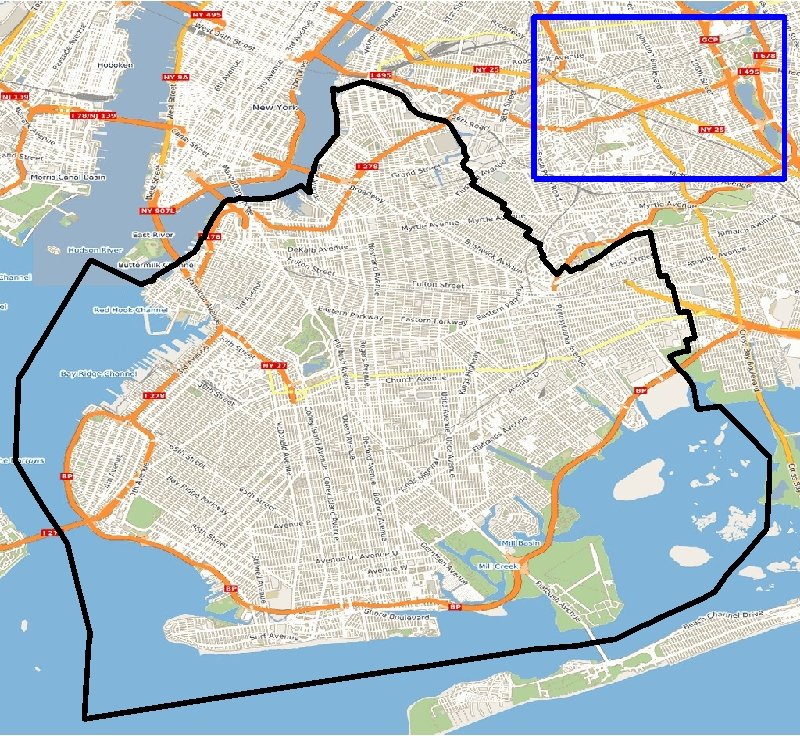}
  
  \caption{A coverage map for the Brooklyn (black) and Queens (blue)
  regions in our dataset.}

  \label{fig:s_coverage}
\end{figure*}

\begin{figure*}
  
  \newlength{\piemargin}
  \setlength{\piemargin}{4pt}

  \centering
 
  \begin{subfigure}{1\linewidth}
    \centering
    \includegraphics[height=.2\linewidth,trim={\piemargin, 1cm, \piemargin, \piemargin},clip]{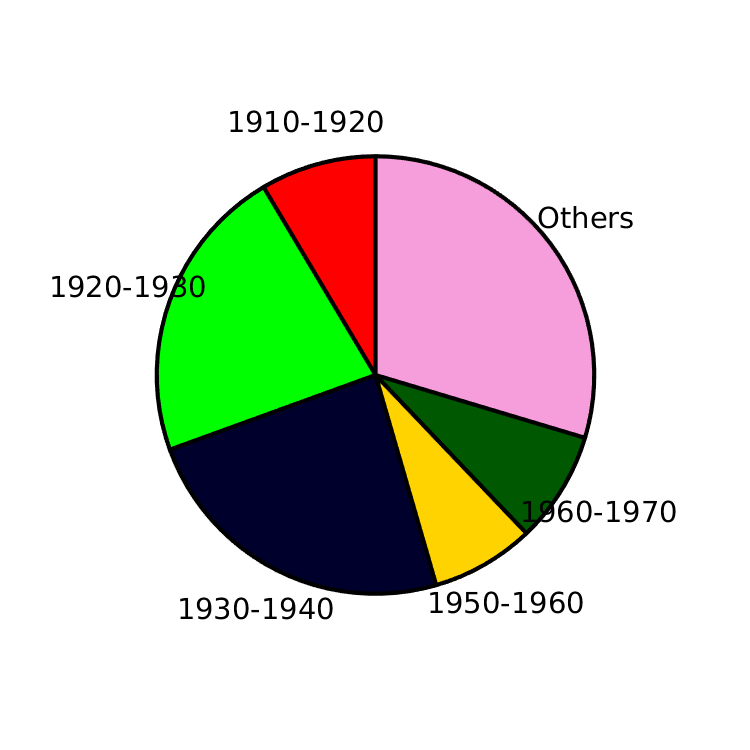}
    \includegraphics[height=.2\linewidth,trim={\piemargin, 1cm, \piemargin, \piemargin},clip]{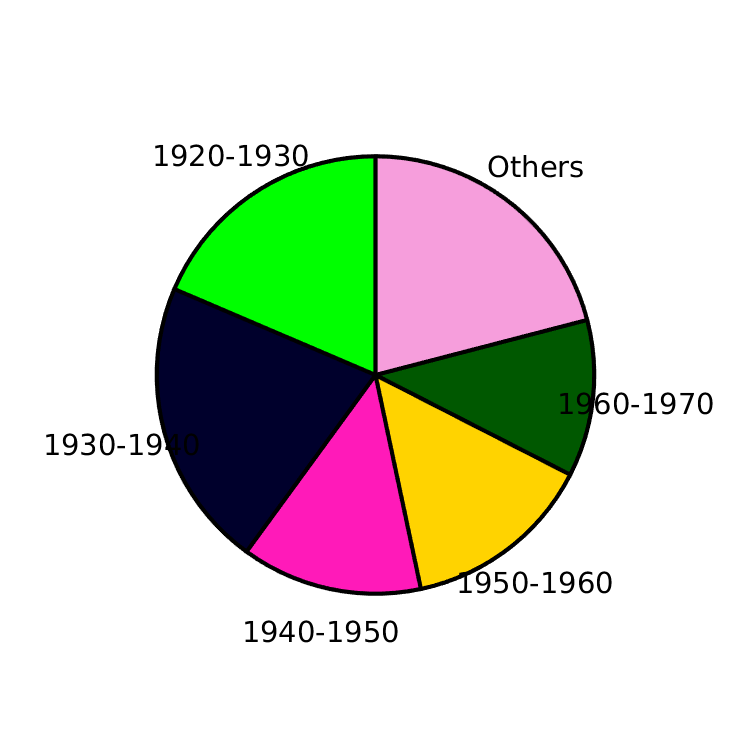}
    \caption{Age}
  \end{subfigure}

  \begin{subfigure}{1\linewidth}
    \centering
    \includegraphics[height=.2\linewidth,trim={\piemargin, 1cm, \piemargin, \piemargin},clip]{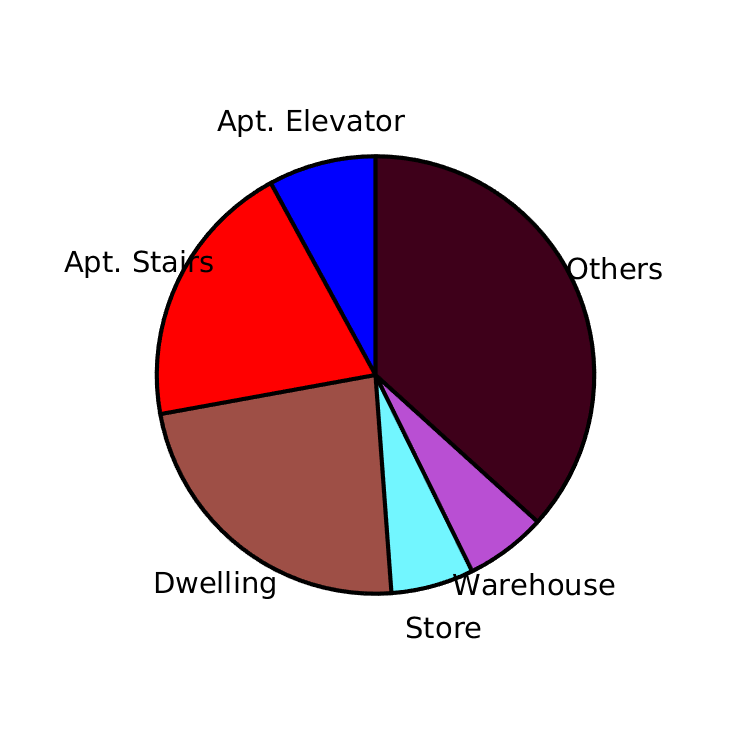}
    \includegraphics[height=.2\linewidth,trim={\piemargin, 1cm, \piemargin, \piemargin},clip]{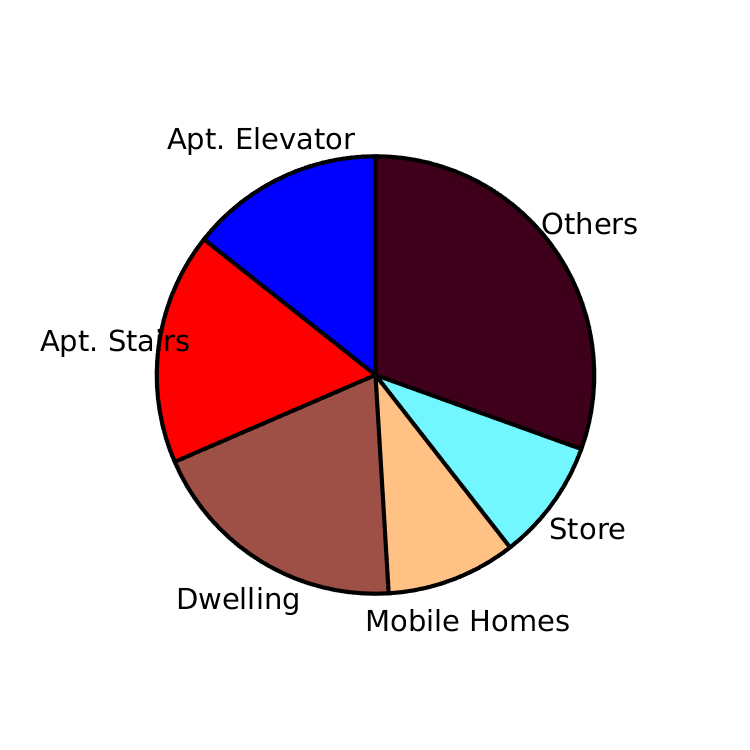}
    \caption{Function}
  \end{subfigure}
  
  \begin{subfigure}{1\linewidth}
    \centering
    \includegraphics[height=.2\linewidth,trim={\piemargin, 1cm, \piemargin, \piemargin},clip]{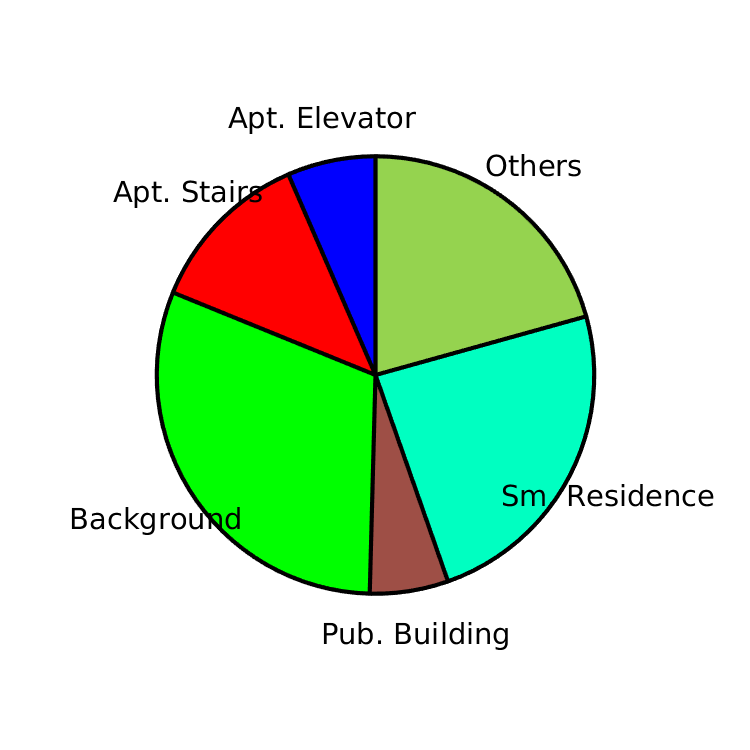}
    \includegraphics[height=.2\linewidth,trim={\piemargin, 1cm, \piemargin, \piemargin},clip]{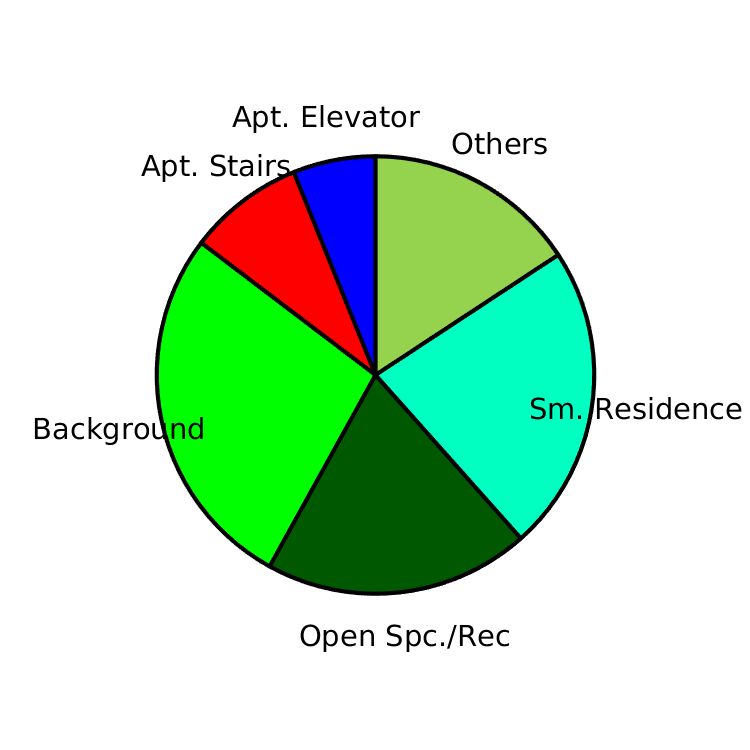}
    \caption{Land Use}
  \end{subfigure}
  
  \caption{Distribution of labels for the Brooklyn (left) and Queens
  (right) test sets.}

  \label{fig:s_data-dist}
\end{figure*}

\figref{s_coverage} shows the spatial coverage of the Brooklyn and
Queens regions in our dataset. \figref{s_data-dist} visualizes the label
distributions for the Brooklyn and Queens test sets. Compared to
Brooklyn, Queens has significantly different label occurrence. For
example, for land use classification, Brooklyn has more ``Public
Buildings'', while Queens has more ``Open Space/Recreation''.

\section{Adaptive Bandwidth Visualization}

\begin{figure*}

  \centering
  
  \begin{subfigure}{.49\linewidth}
    \centering
    \includegraphics[width=1\linewidth]{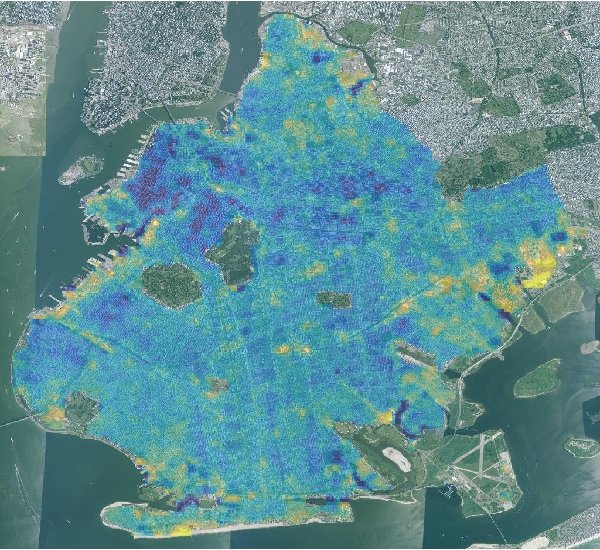}
    \caption{Brooklyn}
  \end{subfigure}
  \begin{subfigure}{.49\linewidth}
    \centering
    \includegraphics[width=1\linewidth]{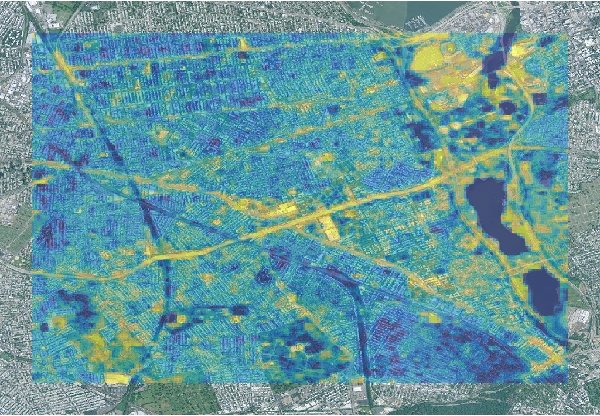}
    \caption{Queens}
  \end{subfigure}
  
  \caption{Adaptive kernel bandwidth estimation. For each location we
  show the mean of the estimated optimal kernel bandwidth parameters,
  for the task of land use classification, computed using the {\em
  unified (adaptive)} method.} 

  \label{fig:s_adaptive}
\end{figure*}

In \figref{s_adaptive} we visualize the estimated kernel bandwidth
parameters, computed using our {\em unified (adaptive)} method for
the task of land use classification, as a map for the Brooklyn and
Queens regions. For each location, we display the mean of the diagonal
entries of the kernel bandwidth matrix, $\Sigma$. These results show
that the adaptive method is adjusting based on the underlying terrain.

\section{Semantic Segmentation Results}

\begin{figure*}

  \centering
  
  \includegraphics[width=.3\linewidth,trim={1.45cm 1cm 3.5cm .1cm},clip]{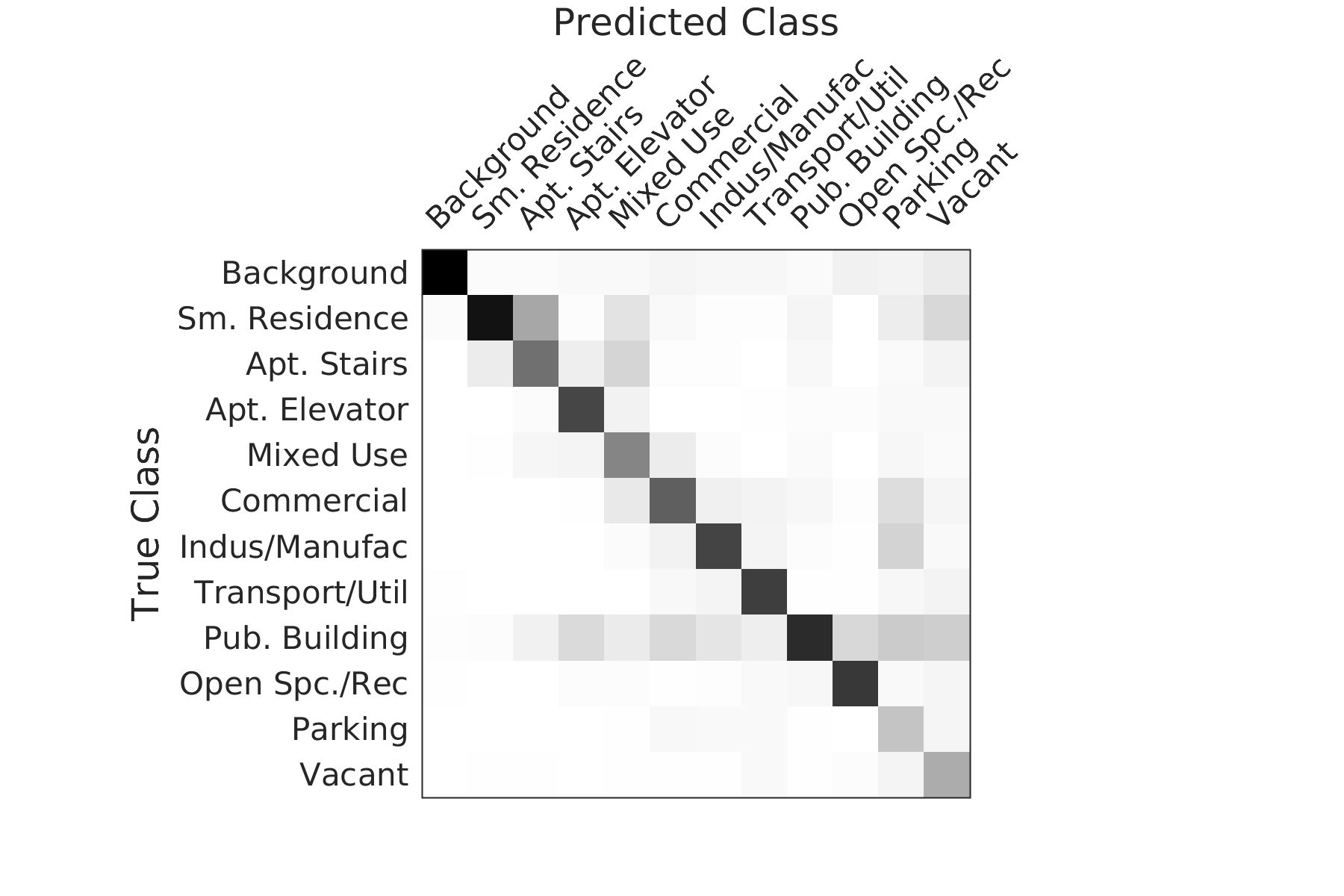}
  \includegraphics[width=.3\linewidth,trim={2.1cm 1cm 2.7cm 0cm},clip]{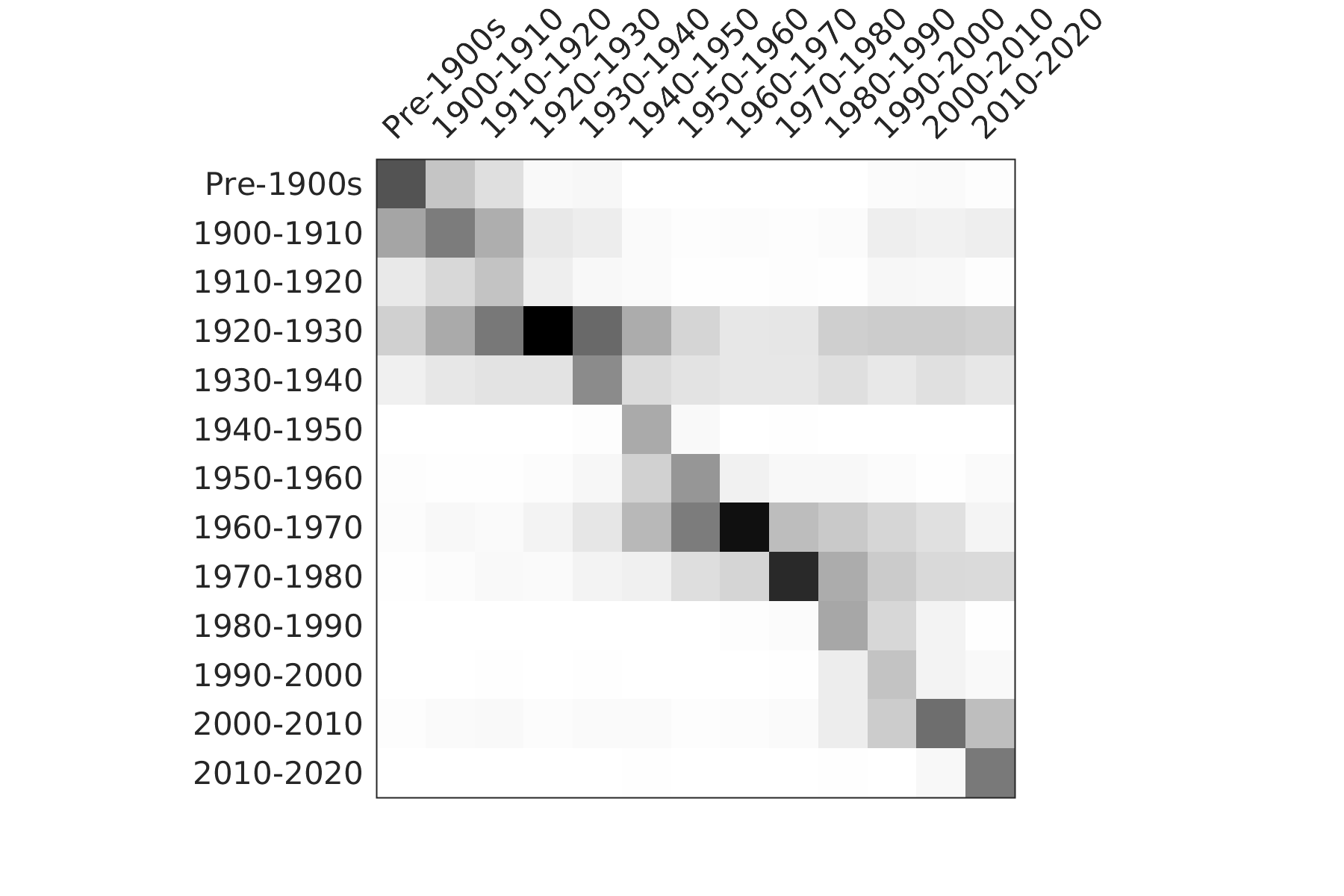}
  \includegraphics[width=.3\linewidth,trim={1.8cm 1cm 2.8cm 0cm},clip]{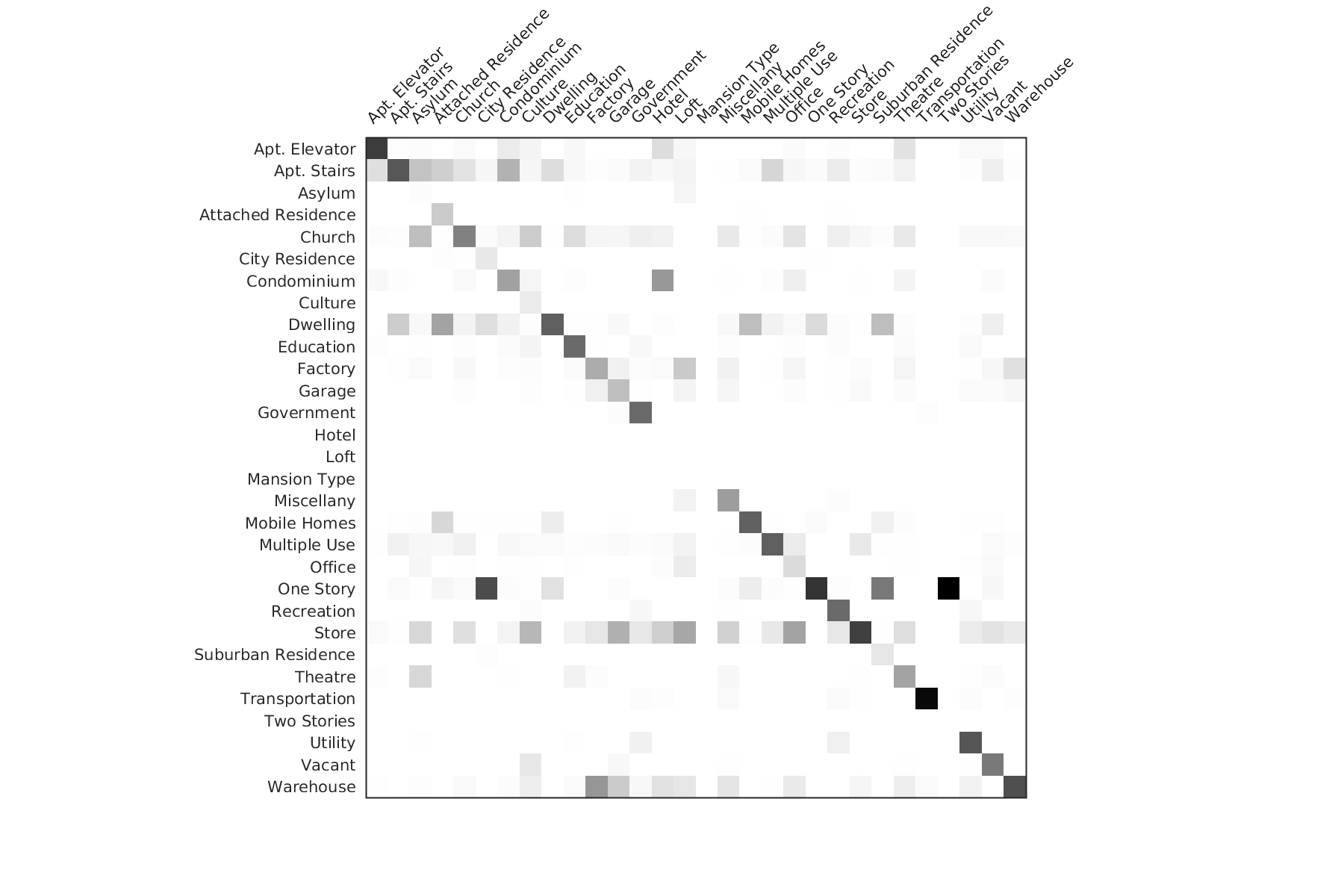}

  \caption{Confusion matrices for classifying land use (left),
  estimating building age (middle), and identifying building function
  (right). These results were computed using our {\em unified
  (adaptive)} approach for the Brooklyn test region.}

  \label{fig:s_confusion}
\end{figure*}

\figref{s_confusion} shows confusion matrices for all three labeling
tasks we consider (land use, age, function), each computed using the
{\em unified (adaptive)} approach, for the Brooklyn test set. For
building function estimation, we aggregate the 206 building classes
into 30 higher-level classes.  Classes are merged according to a
hierarchy outlined by the New York City Department of City Planning in
the PLUTO dataset. Despite the challenging nature of these tasks, our
method seems to make sensible mistakes. For example, for the task of
estimating building age, nearby decades are most often confused.

We report performance, top-1 accuracy and mean region intersection
over union (mIOU), for building function estimation after aggregating
the classes.  For {\em unified (adaptive)}, on the Brooklyn test set,
top-1 accuracy increases to $61.08\%$ and mIOU increases to $30.40\%$.
Similarly for Queens, top-1 accuracy increases to $52.01\%$ and mIOU
increases to $14.99\%$.

\begin{figure*}

  \centering
  
  \includegraphics[width=.3\linewidth]{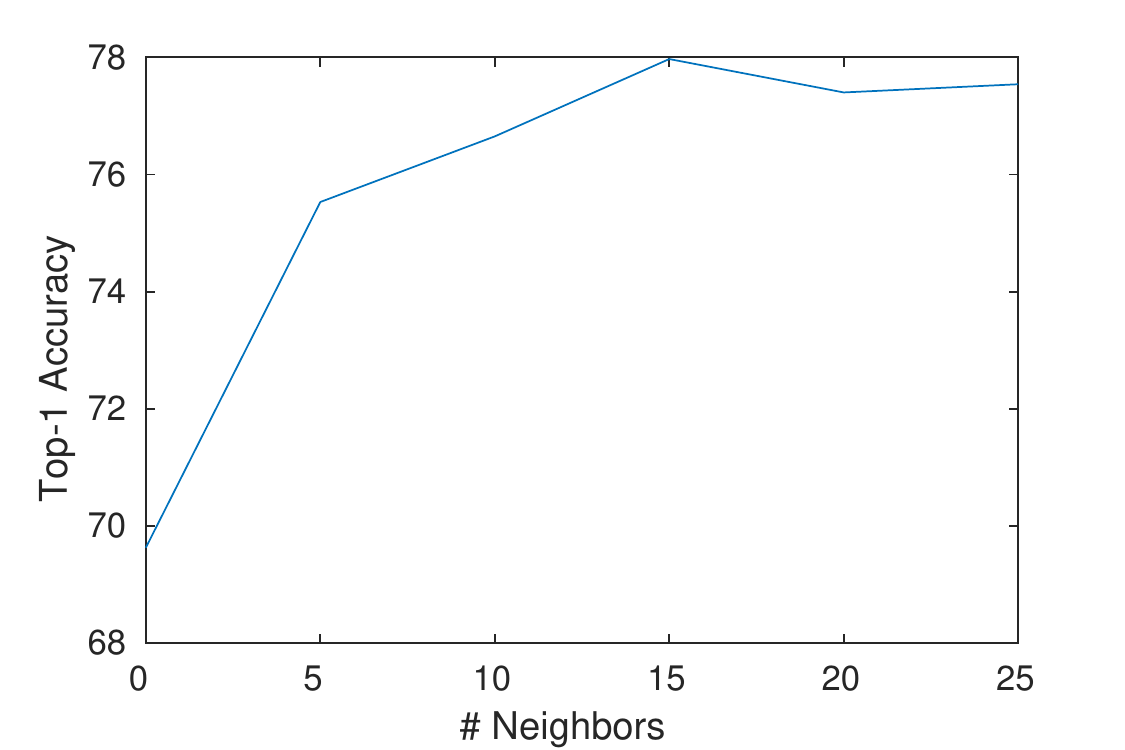}
  
  \caption{Varying the number of nearby ground-level images (land use
  classification). Each point corresponds to an instance of our {\em
  unified (adaptive)} method, except $N=0$ which reflects the
  performance of the {\em remote} baseline.}

  \label{fig:s_nearby}
\end{figure*}

In our experiments, we considered the $N=20$ closest ground-level
images, chosen empirically based on available computational resources.
Theoretically, there is no downside to including as many ground-level
images as possible. However, we explored at what point performance
might saturate. We performed this experiment for land use
classification, using our {\em unified (adaptive)} approach, varying
$N$ in increments of 5 up to 25, and found that performance saturated
at $N=15$, but this was just one dataset/task. \figref{s_nearby}
visualizes the results of this experiment using top-1 accuracy. 

For each labeling task, we show additional semantic labeling results
in \figref{s_labeling}.

\clearpage

\begin{figure*}

  \centering

  \setlength\tabcolsep{1pt}

  \begin{tabular}{lcccccc}
    \raisebox{.25\height}{\rotatebox{90}{ground truth}} &
    \includegraphics[width=.16\linewidth]{{{results/supplemental/landuse/154325_197258}}} &
    \includegraphics[width=.16\linewidth]{{{results/supplemental/landuse/154329_197326}}} &
    \includegraphics[width=.16\linewidth]{{{results/supplemental/landuse/154361_197279}}} &
    \includegraphics[width=.16\linewidth]{{{results/supplemental/landuse/154355_197262}}} &
    \includegraphics[width=.16\linewidth]{{{results/supplemental/landuse/154351_197276}}} &
    \includegraphics[width=.16\linewidth]{{{results/supplemental/landuse/154354_197257}}} \\
    
    \raisebox{.06\height}{\rotatebox{90}{\em unified (adaptive)}} &
    \includegraphics[width=.16\linewidth]{{{results/supplemental/landuse/154325_197258_adaptive}}} &
    \includegraphics[width=.16\linewidth]{{{results/supplemental/landuse/154329_197326_adaptive}}} &
    \includegraphics[width=.16\linewidth]{{{results/supplemental/landuse/154361_197279_adaptive}}} &
    \includegraphics[width=.16\linewidth]{{{results/supplemental/landuse/154355_197262_adaptive}}} &
    \includegraphics[width=.16\linewidth]{{{results/supplemental/landuse/154351_197276_adaptive}}} &
    \includegraphics[width=.16\linewidth]{{{results/supplemental/landuse/154354_197257_adaptive}}} \\
    \vspace{1cm} 
  \end{tabular}
  \begin{tabular}{lcccccc}
    \raisebox{.25\height}{\rotatebox{90}{ground truth}} &
    \includegraphics[width=.16\linewidth]{{{results/supplemental/age/154419_197140}}} &
    \includegraphics[width=.16\linewidth]{{{results/supplemental/age/154419_197247}}} &
    \includegraphics[width=.16\linewidth]{{{results/supplemental/age/154430_197103}}} &
    \includegraphics[width=.16\linewidth]{{{results/supplemental/age/154430_197225}}} &
    \includegraphics[width=.16\linewidth]{{{results/supplemental/age/154437_197176}}} &
    \includegraphics[width=.16\linewidth]{{{results/supplemental/age/154433_197174}}} \\
    
    \raisebox{.06\height}{\rotatebox{90}{\em unified (adaptive)}} &
    \includegraphics[width=.16\linewidth]{{{results/supplemental/age/154419_197140_adaptive}}} &
    \includegraphics[width=.16\linewidth]{{{results/supplemental/age/154419_197247_adaptive}}} &
    \includegraphics[width=.16\linewidth]{{{results/supplemental/age/154430_197103_adaptive}}} &
    \includegraphics[width=.16\linewidth]{{{results/supplemental/age/154430_197225_adaptive}}} &
    \includegraphics[width=.16\linewidth]{{{results/supplemental/age/154437_197176_adaptive}}} &
    \includegraphics[width=.16\linewidth]{{{results/supplemental/age/154433_197174_adaptive}}} \\
    \vspace{1cm} 
  \end{tabular}
  \begin{tabular}{lcccccc}
    \raisebox{.25\height}{\rotatebox{90}{ground truth}} &
    \includegraphics[width=.16\linewidth]{{{results/supplemental/function/154318_197280}}} &
    \includegraphics[width=.16\linewidth]{{{results/supplemental/function/154325_197300}}} &
    \includegraphics[width=.16\linewidth]{{{results/supplemental/function/154329_197326}}} &
    \includegraphics[width=.16\linewidth]{{{results/supplemental/function/154326_197300}}} &
    \includegraphics[width=.16\linewidth]{{{results/supplemental/function/154345_197259}}} &
    \includegraphics[width=.16\linewidth]{{{results/supplemental/function/154374_197282}}} \\
    
    \raisebox{.06\height}{\rotatebox{90}{\em unified (adaptive)}} &
    \includegraphics[width=.16\linewidth]{{{results/supplemental/function/154318_197280_adaptive}}} &
    \includegraphics[width=.16\linewidth]{{{results/supplemental/function/154325_197300_adaptive}}} &
    \includegraphics[width=.16\linewidth]{{{results/supplemental/function/154329_197326_adaptive}}} &
    \includegraphics[width=.16\linewidth]{{{results/supplemental/function/154326_197300_adaptive}}} &
    \includegraphics[width=.16\linewidth]{{{results/supplemental/function/154345_197259_adaptive}}} &
    \includegraphics[width=.16\linewidth]{{{results/supplemental/function/154374_197282_adaptive}}} \\
  \end{tabular}
  
  \caption{Additional semantic labeling results for classifying land
  use (top), estimating build age (middle) and identifying building
  function (bottom).}

  \label{fig:s_labeling}

\end{figure*}

\end{document}